\documentclass[11pt,a4paper]{article}
\usepackage[hyperref]{acl2020}
\usepackage{times}
\usepackage{latexsym}
\usepackage{comment}
\usepackage{todonotes}
\usepackage{caption}
\usepackage{subcaption}
\usepackage{booktabs}
\usepackage{siunitx}
\usepackage{tabularx}

\graphicspath{{./}{./figures/}}

\definecolor{customgreen}{rgb}{0.0, 0.4, 0.0}

\usepackage{microtype}

\aclfinalcopy 

\title{Quantifying the Effects of COVID-19 on Mental Health Support Forums}

\author{\parbox{\linewidth}{\centering Laura Biester\thanks{\ \ Denotes equal contribution.}\ , Katie Matton\footnotemark[1]\ , Janarthanan Rajendran, \\Emily Mower Provost, Rada Mihalcea}\\
Computer Science \& Engineering, University of Michigan, USA\\
\texttt{\{lbiester,katiemat,rjana,emilykmp,mihalcea\}@umich.edu}}

\date{}

\begin{document}
\maketitle

\begin{abstract}
The COVID-19 pandemic, like many of the disease outbreaks that have preceded it, is likely to have a profound effect on mental health. Understanding its impact can inform strategies for mitigating negative consequences. In this work, we seek to better understand the effects of COVID-19 on mental health by examining discussions within mental health support communities on Reddit. First, we quantify the rate at which COVID-19 is discussed in each community, or subreddit, in order to understand levels of preoccupation with the pandemic. Next, we examine the volume of activity in order to determine whether the quantity of people seeking online mental health support has risen. Finally, we analyze how COVID-19 has influenced language use and topics of discussion within each subreddit.


%
\end{abstract}

\section{Introduction}
The implications of COVID-19 extend far beyond its immediate physical health effects. Uncertainty and fear surrounding the disease and its effects, in addition to a lack of consistent and reliable information, contribute to rising levels of anxiety and stress \cite{torales2020outbreak}. Policies designed to help contain the disease also have significant consequences. Social distancing policies and lockdowns lead to increased feelings of isolation and uncertainty \cite{Huremovic2019}. They have also triggered an economic downturn \cite{Sahin2020}, resulting in soaring unemployment rates and causing many to experience financial stress. Therefore, in addition to the profound effects on physical health around the world, psychiatrists have warned that we should also brace for a mental health crisis as a result of the pandemic \cite{qiu2020nationwide,greenberg2020managing,yao2020patients,torales2020outbreak}.

Indeed, the literature on the impact of past epidemics indicates that they are associated with a myriad of adverse mental health effects. In a review of studies on the 2002-2003 SARS outbreak, the 2009 H1N1 influenza outbreak, and the 2018 Ebola outbreak, \citet{Chew2020} found that anxiety, fear, depression, anger, guilt, grief, and post-traumatic stress were all commonly observed psychological responses. Furthermore, many of the factors commonly cited for inducing these responses are applicable to the situation with COVID-19; these include: fear of contracting the disease, a disruption in daily routines, isolation related to being quarantined, and uncertainty regarding the disease treatment process and outcomes, the well being of loved ones, and one's economic situation. 
	
While disease outbreaks pose a risk to the mental health of the general population, research suggests that this risk is heightened for those with pre-existing mental health concerns. People with mental health disorders are particularly susceptible to experiencing negative mental health consequences during times of social isolation \cite{usher2020life}. Further, as \citet{yao2020patients} warns, they are likely to have a stronger emotional response to the feelings of fear, anxiety, and depression that come along with COVID-19 than the general population.
	
Given the potential for the COVID-19 outbreak to have devastating consequences for mental health, it is critical that we work to understand and mitigate its negative psychological effects. In this work, we use Reddit, a popular social media platform, to study how COVID-19 has impacted the behavior of people who express mental health concerns. We focus on three Reddit sub-forums, referred to as subreddits, that are designed to offer peer support for users who are struggling with specific types of mental illness. We first measure changes in the amount of activity on these subreddits to determine whether, for a subset of the population, the need for online mental health support has increased or decreased. We then analyze the content of subreddit discussions, gaining insight into how the pandemic affects what people choose to discuss when seeking support and what types of issues push people to seek support. Our findings provide insights into how and when COVID-19 related stressors affect people dealing with mental health concerns. Critically, we believe that this information can help to better understand the nature of a potential COVID-19 related mental health crisis.

\section{Related Work}
\subsection{Studying Mental Health via Social Media}
In the past decade, social media has emerged as a powerful tool for understanding human behavior, and correspondingly mental health. A growing number of studies have applied computational methods to data collected from social media platforms in order to characterize behavior associated with mental health illnesses and to detect and forecast mental health outcomes (see \citet{chancellor2020methods} for a comprehensive review). 

Reddit is a particularly well-suited platform for studying mental health due to its semi-anonymous nature, which  encourages user honesty and reduces inhibitions associated with self-disclosure 
\cite{de2014mental}. Additionally, Reddit contains subreddits that act as mental health support forums (e.g., r/Anxiety, r/depression, r/SuicideWatch), which enable a more targeted analysis of users experiencing different mental health conditions. A number of existing works have focused on characterizing patterns of discourse within these mental health communities on Reddit. These include studies that have analyzed longitudinal trends in topic usage and word choice \cite{chakravorti2018detecting}, the relationship between user participation styles and topic usage \cite{feldhege2020says}, and the discourse patterns specific to self-disclosure, social support, and anonymous posting \cite{pavalanathan2015identity, de2014mental}.

Other studies of Reddit mental health communities have aimed to quantify and forecast changes in user behavior. \citet{de2016discovering} presented a model for predicting the likelihood that users transition from discussing mental health generally to engaging in suicidal ideation. \citet{li2018text} analyzed linguistic style measures associated with increasing vs decreasing participation in mental health subreddits over the course of a year. \citet{Kumar2015} examined how posting activity in r/SuicideWatch changes following a celebrity suicide. Our work similarly focuses on analyzing temporal patterns in user activity, but we aim to characterize changes associated with COVID-19.

\subsection{Mental Health and COVID-19}

Since the first cases of COVID-19 were reported in December 2019, there have been a few preliminary studies of its impact on mental health. In a survey of the general public of China during the initial outbreak, a majority of respondents perceived the psychological impact of the outbreak to be moderate-to-severe and about one-third reported experiencing moderate-to-severe anxiety \cite{wang2020immediate}. Studies of the impact of COVID-19 among residents of Liaoning Province, China \cite{zhang2020impact} and the adult Indian population \cite{roy2020study} also found notable rates of mental distress. 

There is a small set of studies that have examined the mental health consequences of COVID-19 by analyzing online behaviors. \citet{jacobson2020flattening} explored the short-term impact of stay-at-home orders in the United States by analyzing changes in the rates of mental health-related Google search queries immediately after orders were issued. Their results showed that rates of mental health queries increased leading up to the issuance of stay-at-home-orders, but then plateaued after they went into effect; however they did not consider the longer-term implications of the stay-at-home orders on mental health. \citet{li2020impact} measured psycholinguistic attributes of posts on Weibo, a Chinese social media platform, before and after the Chinese National Health Commission declared COVID-19 to be an epidemic. Their findings showed that expressions of negative emotions and sensitivity to social risks increased following the declaration. 
\citet{wolohan2020estimating} used a LSTM model to classify depression among Reddit users in April 2020, finding a higher than normal depression rate.

Our work similarly aims to measure changes in online behavior as a means of understanding the relationship between COVID-19 and mental health. However, two notable differences are: (1) instead of analyzing the short-term impact of a specific COVID-related event, we examine more general changes that have occurred during a three-month period of the outbreak; and (2) we focus our analysis on activity within mental health forums, which allows us to examine the impact of COVID-19 specifically on individuals who have expressed mental health concerns.


\section{Data}
We collect Reddit posts from three mental health subreddits using the Pushshift API \footnote{As with other social media datasets, there may be noise in the form of API changes and data removed after collection. For the dates involved in our study, static Pushshift dump files were not yet available.} \cite{Baumgartner2020}: r/Anxiety, r/depression, and r/SuicideWatch, from January 2017 to May 2020. The reasons for analyzing these three subreddits are twofold: first, over the three and a half years represented in our data, these subreddits have a significant amount of activity ($\geq 50$ posts almost every day), making it feasible to treat daily values as a time series. Second, because the subreddits provide support for different mental health disorders, their users may have been affected differently by COVID-19. We separate the data into two time periods: pre-COVID (January 1, 2017 - February 29, 2020) and post-COVID (March 1, 2020 - May 31, 2020), roughly delineating when COVID-19 began to have a serious impact on those in the United States, where the majority of Reddit users are concentrated.\footnote{https://www.alexa.com/siteinfo/reddit.com} This choice of dates was informed by our analysis of the rates at which COVID-19 related words were discussed in each subreddit (see Section~\ref{sec:covid-discussion}), which we found hovered around 0-5\% before rising sharply near the beginning of March. 


We exclude posts 
where the author or text is marked as `[removed]' or `[deleted]'. 
Table \ref{tab:data-stats} shows the average number of daily posts for r/Anxiety, r/depression, and r/SuicideWatch.

\begin{table}
\centering
\resizebox{\linewidth}{!}{
\begin{tabular}{@{}llll@{}}
\toprule
       & r/Anxiety & r/depression & r/SuicideWatch \\ \midrule
2017   & 95                    & 279         & 91            \\
2018   & 164          & 449               & 188                \\
2019   & 211               & 622              & 285                  \\ 
2020  & 243 & 618 & 370 \\\bottomrule
\end{tabular}}
\caption{Average number of posts per day across the three subreddits in our dataset.}
\label{tab:data-stats}
\end{table}

\section{Methodology}
\subsection{Reddit Activity Metrics}\label{sec:metrics}
We begin by creating a lexicon of words that are commonly used to refer to COVID-19. This allows us to determine the extent to which users in each subreddit are discussing COVID-19, and also gives us a clearer idea of when COVID-19 began to directly affect discussion in the mental health subreddits. We based the lexicon on fifteen twitter search keywords from \citet{huang_xiaolei_2020_3735015}, and added six additional words that we believed would be indicative of discussion about COVID-19.\footnote{corona, outbreak, pandemic, rona, sars-cov-2, virus}

To study changes in the number of users seeking mental health support in subreddits, we record the author usernames for each post in our dataset. We note that, since individuals can create multiple accounts under different usernames, the number of unique usernames associated with posts is likely not equal to the true number of unique users; however, we believe it is a reasonable proxy.

To study changes in content that occur during the pandemic, we use the Linguistic Inquiry and Word Count (LIWC) lexicon \cite{pennebaker2015development} and Latent Dirichlet Allocation (LDA) topic modeling \cite{blei2003latent}. The LIWC lexicon consists of seventy-three hierarchical psycholinguistic word categories, encapsulating properties including linguistic categories (e.g. 1st person plural pronouns, verbs), emotions (e.g. anxiety, sadness), time (e.g. present, future), and personal concerns (e.g. work, money, death). To capture the discussion topics that are common in the r/Anxiety, r/depression, and r/SuicideWatch subreddits specifically, we train a topic model on posts from these subreddits. We ensure that discussions from each of the subreddits are equally represented in our training dataset by downsampling the posts from the subreddits with more data. We use the implementation of LDA topic modeling provided in the MALLET toolkit \cite{mallet} and train models with $k=5,10,..,40$ topics. We select a single model to use in our analysis by examining their coherence scores, a measure of the semantic similarity of high probability words within each topic \cite{mimno2011optimizing}. As coherence tends to increase with increasing $k$, we select $k$ as the first local maxima of coherence scores, which we found to be $k=25$.  

In Appendix~\ref{sec:topics-keywords}, we show the 25 topics obtained from our topic model, along with the highest probability words associated with each topic. We also provide labels that summarize the essence of each topic, which we created by examining their representative words. Common themes of discussion include: daily life concerns (e.g. school, work, sleep and routine), personal relationships (e.g. friends, family, relationships), and mental health struggles (e.g. anxiety, suicide, medical treatment).

When using text from posts, we remove special characters and sequences, such as newlines, quotes, emails, and tables. To represent the text of a post, we concatenate the title with the text content, as was done in prior work \cite{chakravorti2018detecting}. We apply additional pre-processing steps for our topic modeling analysis: (1) we remove a set of common stopwords that do not appear in the LIWC dictionary (we kept these as they have been found to have psychological meaning), (2) we form bigrams from pairs of words that commonly appear together, and (3) we lemmatize each word.

\subsection{Time Series Analysis}\label{sec:time-series-analysis}
We treat the task of identifying changes in subreddit activity patterns as a time series intervention analysis problem. Our basic approach involves: (1) fitting a time series model to the pre-COVID observations for each of the metrics described above and then (2) examining how the values forecasted by the model compare to the observed values during the post-COVID time period. It is worth noting that the one existing study we found examining the impact of an event on activity within mental health subreddits employs a different approach: they use a t-test to compare the observations from ``before" vs``after" the event \cite{Kumar2015}. However, their problem setup differs from ours in that they consider a much shorter period of time (4 weeks total), so the effects of seasonality and longer-term trends are likely reduced. In contrast, we find that there is often a strong trend over time and seasonal component in our data, making a direct comparison of two time periods with a t-test unreliable.


We smooth each time series and remove day-of-week related fluctuations by computing a seven-day rolling mean over the time series. We use the Prophet model \cite{Taylor2018} to create a model of the period before COVID-19. This model was initially created by Facebook to forecast time series on their platform, such as the number of events created per day or the number of active users; we find that our time series, also compiled from social media, have many similar properties. The Prophet model is an additive regression model with three components:

\begin{equation}
    y(t) = g(t) + s(t) + h(t) + \epsilon_t
\end{equation}

The trend is encapsulated by $g(t)$, a piecewise linear model. The seasonality of the data is captured by $s(t)$, which is approximated using a Fourier series; as we smooth our data on a weekly basis, we utilize only yearly seasonality, excluding the optional weekly and daily seasonality components. The third term, $h(t)$, represents holidays; we find that adding a list of US holidays reduces error for most our our time series in the pre-COVID period, likely because the Reddit population is centered in the United States. Finally, $\epsilon_t$ represents the error, in this case fluctuations in the time series that are not captured by the model.

After training the model on the pre-COVID data, we predict values for the post-COVID period. If we assume that there is no change during this time period, we would expect the predicted values to be near the true values, given that the model does a good job fitting the trend and seasonal components. The model computes uncertainty intervals over the predicted values by simulating ways in which the trend may change during the period of the forecast; we use this method to compute the 95\% prediction interval. Our null hypothesis is that there has been no change in trend; in this case, we would expect 5\% of the data in the post-COVID period to fall outside of the prediction interval. Our alternate hypothesis is that there was a change in the trend of the time series (which may be attributable to COVID-19); in this case, more than 5\% of the data in the post-COVID period will fall outside of the prediction interval. We apply a one-sample proportion test to assess whether the proportion of observations outside of the prediction interval in the post-COVID period is significantly greater than 5\%. The details of this test are in Appendix~\ref{sec:sig-testing}.

\section{Findings}
\subsection{Do people in different mental health subreddits discuss COVID-19 at different rates?}
\label{sec:covid-discussion}
\begin{figure}[ht]
	\centering
		\includegraphics[width=\linewidth]{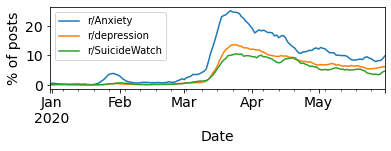}
	\caption{Percent of posts mentioning COVID-19 related words across mental health subreddits.}
	\label{fig:covid19-lexicon}
	\label{fig:covid19-lexicon-post}
\end{figure}

Using our COVID-19 lexicon (Section \ref{sec:metrics}), we compute the percentage of posts per day that mention any words related to COVID-19, as shown in Figure \ref{fig:covid19-lexicon-post}.

When choosing the date to consider as the beginning of the post-COVID period in our time series analysis, we considered March 1st, 2020 as a sensible date, as it aligns with the period at which the United States (where the majority of Reddit users reside) began to take COVID-19 seriously. March 1st directly followed the first announced COVID-19 death on February 28th, 2020, and preceeded state lockdowns and school closures. Our analysis also offers compelling evidence that COVID-19 began to have a serious impact on discussions in all three analyzed subreddits around the beginning of March 2020, as is clear in the spikes in Figure \ref{fig:covid19-lexicon-post}.

Notably, we also see a stark difference in the volume of discussion involving COVID-19 in the three subreddits; in r/Anxiety, discussion of COVID-19 is more frequent than it is in r/depression or r/SuicideWatch, and begins earlier; users who come to Reddit to discuss anxiety disorders clearly began to feel some impact from COVID-19 in late January, when reports of lockdowns in China first appeared in the news. We conclude that the discussions in r/Anxiety are likely to more strongly be affected by COVID-19, and proceed to analyze the nature of those effects.





\subsection{Has COVID-19 changed the number of users seeking support in mental health subreddits?}

\begin{figure}[ht]
	\centering
	\includegraphics[width=\linewidth]{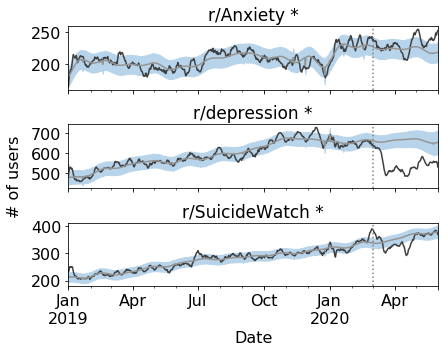}
	\caption{Daily active users over time. The grey line is the Prophet forecast, the shaded area is the 95\% prediction interval, and the black line is the true value. Subreddits marked with * have a statistically significant percentage of outliers ($\alpha = 0.05$).}
	\label{fig:n_users}
\end{figure}
We report the daily number of unique users who post in each subreddit in Figure~\ref{fig:n_users}. We observe a significant increase in the number of users who post in the r/Anxiety subreddit during the post-COVID period. This finding is consistent with prior work that has found that epidemics often lead to increased rates of anxiety \cite{torales2020outbreak}. Meanwhile, in both the r/depression and r/SuicideWatch subreddits, we find that there are significant decreases in the number of users posting. In r/depression, we observe a substantial drop in posting rates around mid-March. Activity in this subreddit remains abnormally low into late-April, when it starts to revert back towards the forecasted values. In r/SuicideWatch, the drop in user activity is less extreme, and we see that the activity levels eventually return to their predicted values. 
Our findings contrast with prior work that found that depressive symptoms are commonly observed during pandemics  \cite{Chew2020}. 
However, research also indicates that delayed depression is a common symptom trajectory following disaster events \cite{Pennebaker93Social,nandi2009patterns}; therefore, future analysis of how these patterns change in the longer-term may help to better understand the current trends.

\subsection{Has COVID-19 led to changes in the discussions users have surrounding mental health?}
To determine what changes have occurred in conversations surrounding mental health, we use two types of features: LIWC categories and topics obtained from an LDA model. The LIWC features give us a better idea of how common language dimensions have changed, while the LDA-derived topics allow us to explore areas of discussion that are typically of concern in these subreddits. For each of the metrics, we examine changes that have occurred since COVID-19 by computing the proportion of outliers produced by our forecasting model (see Section~\ref{sec:time-series-analysis}) in the post-COVID period.

\subsubsection{LIWC Analysis}\label{sec:liwc-results}
\begin{table}
\centering
\resizebox{\linewidth}{!}
{\begin{tabular}{@{}lcc|lcc|lcc@{}}
\toprule
\multicolumn{3}{c|}{r/Anxiety}            & \multicolumn{3}{c|}{r/depression}         & \multicolumn{3}{c}{r/SuicideWatch}       \\ \midrule
Category & \multicolumn{2}{c|}{\% Outliers} & Category & \multicolumn{2}{c|}{\% Outliers}  & Category & \multicolumn{2}{c}{\% Outliers}   \\ \midrule
MOTION\textsuperscript{*}        & 79       & {\color{red}$\downarrow$} & YOU\textsuperscript{*}           & 55       & {\color{red}$\downarrow$} & PREP\textsuperscript{*}          & 33       & {\color{customgreen}$\uparrow$}   \\
WORK\textsuperscript{*}          & 73       & {\color{red}$\downarrow$} & CONJ\textsuperscript{*}          & 51       & {\color{red}$\downarrow$} & SPACE\textsuperscript{*}         & 33       & {\color{customgreen}$\uparrow$}   \\
I\textsuperscript{*}             & 68       & {\color{red}$\downarrow$} & MOTION\textsuperscript{*}        & 45       & {\color{red}$\downarrow$} & NETSPEAK\textsuperscript{*}      & 23       & {\color{customgreen}$\uparrow$}   \\
BODY\textsuperscript{*}          & 61       & {\color{customgreen}$\uparrow$}   & QUANT\textsuperscript{*}         & 43       & {\color{customgreen}$\uparrow$}   & ASSENT\textsuperscript{*}        & 23       & {\color{red}$\downarrow$} \\
PPRON\textsuperscript{*}         & 54       & {\color{red}$\downarrow$} & FAMILY\textsuperscript{*}       & 40       & {\color{customgreen}$\uparrow$}   & INFORMAL\textsuperscript{*}      & 22       & {\color{customgreen}$\uparrow$}   \\
RELATIV\textsuperscript{*}       & 54       & {\color{red}$\downarrow$} & ARTICLE\textsuperscript{*}       & 39       & {\color{red}$\downarrow$} & CAUSE\textsuperscript{*}        & 20       & {\color{customgreen}$\uparrow$}   \\
WE\textsuperscript{*}            & 50       & {\color{customgreen}$\uparrow$}   & PRONOUN\textsuperscript{*}      & 38       & {\color{customgreen}$\uparrow$}   & AFFILIATION  & 17       & {\color{red}$\downarrow$} \\
BIO\textsuperscript{*}          & 49       & {\color{customgreen}$\uparrow$}   & REWARD\textsuperscript{*}        & 36       & {\color{red}$\downarrow$} & FOCUSFUTURE   & 16       & {\color{red}$\downarrow$} \\
PERCEPT\textsuperscript{*}       & 42       & {\color{customgreen}$\uparrow$}   & FEEL\textsuperscript{*}          & 35       & {\color{red}$\downarrow$} & NEGEMO        & 15       & {\color{customgreen}$\uparrow$}   \\
CERTAIN\textsuperscript{*}       & 41       & {\color{customgreen}$\uparrow$}   & FOCUSPAST \textsuperscript{*}    & 33       & {\color{customgreen}$\uparrow$}   & CONJ          & 15       & {\color{red}$\downarrow$} \\ 
\bottomrule
\end{tabular}}
\caption{Ten LIWC categories with the highest proportion of outliers in each subreddit. Arrows mark the direction in which the mean of the outliers shifted from the predicted mean. Categories marked with * have a statistically significant percentage of outliers ($\alpha = 0.05$ before Bonferroni correction).}
\label{tab:liwc-results}
\vspace{-3pt}
\end{table}

\begin{figure*}[htp]
	\centering
		\begin{subfigure}{.35\textwidth}
		\centering
		\includegraphics[width=\linewidth]{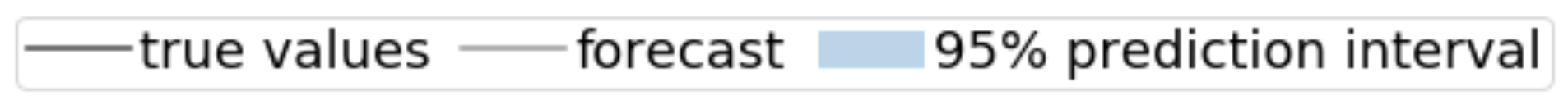}
		\label{fig:legend_users_posts_comments}
	\end{subfigure}
	
	\vspace{-10pt}
	\centering
	\begin{subfigure}[b]{.33\textwidth}
		\centering
		\includegraphics[width=\linewidth]{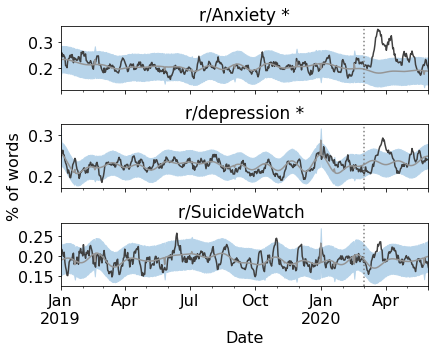}
		\caption{{\sc We} category}\label{fig:liwc-WE}		
	\end{subfigure}%
	\begin{subfigure}[b]{.33\textwidth}
		\centering
		\includegraphics[width=\linewidth]{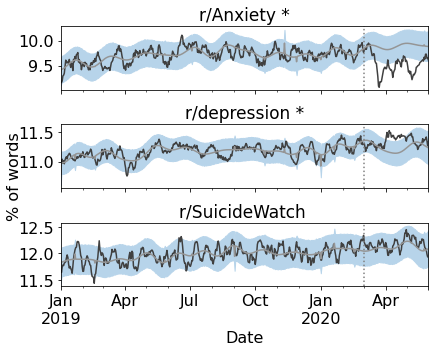}
		\caption{{\sc I} category}\label{fig:liwc-I}
	\end{subfigure}%
	\begin{subfigure}[b]{.33\textwidth}
		\centering
		\includegraphics[width=\linewidth]{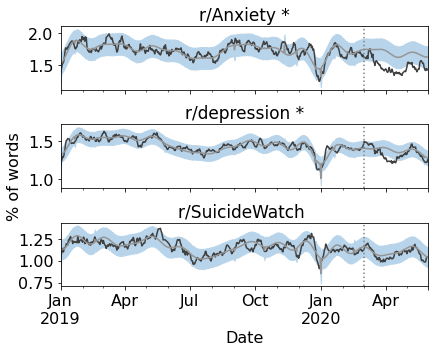}
		\caption{{\sc Work} category}\label{fig:liwc-WORK}
	\end{subfigure}
	\caption{Average daily percent of words across posts from a selection LIWC categories over time. The grey line is the Prophet forecast, the shaded area is the 95\% prediction interval, and the black line is the true value. Subreddits marked with * have a statistically significant percentage of outliers ($\alpha = 0.05$ before Bonferroni correction).}\label{fig:liwc-timeseries}
\end{figure*}

In Table \ref{tab:liwc-results}, we show the ten LIWC categories with the most outliers (outside of the 95\% prediction interval) from March to May of 2020. We observe a lack of consistency between the subreddits, both in the number of outliers and the outliers themselves.

One category that decreases significantly across r/Anxiety and r/depression is {\sc Motion}; this makes intuitive sense as people are traveling and moving around far less. Categories such as {\sc Bio} and {\sc Body} tend to increase in r/Anxiety, where people may fret more over the direct implications of COVID-19 on their physical health; however, this pattern is not present in other subreddits, suggesting that the posts may focus less on the physical health implications. 
We also see consistent changes in time orientation (e.g. {\sc FocusPast}, {\sc FocusFuture}) across subreddits; a higher focus on the past in r/depression, and a lower focus on the future in r/SuicideWatch. While it is not among the categories with the most outliers, there is a statistically significant drop in {\sc FocusFuture} on r/Anxiety and r/depression, indicating that users are less inclined to speak about their concerns for the future in light of the more pressing current concerns.

We also see changes in pronoun usage; the most notable and consistent change across the subreddits is that the usage of {\sc We} increases significantly, especially in the early period of COVID-19 (Figure \ref{fig:liwc-WE}). This indicates a general feeling of community and togetherness, which speaks positively to the support that those in these mental health communities are getting during the pandemic. Similarly, a study on the effects of COVID-19 on mental wellbeing in China found that participants received increased support from friends and family during the pandemic \cite{zhang2020impact}, and seeking social support was listed as a common coping strategy during infections disease outbreaks by \citet{Chew2020}. Simply by posting on Reddit, users are seeking social support, but the increase in first person plural pronouns indicates that they likely are also getting support elsewhere in life. An increase in ``we'' is not specific to mental health communities; researchers have found increases in usage of the pronoun during the early stages of COVID-19 on other subreddits \cite{pandemicprojectblog}.

While there is a significant decrease in {\sc I} words in r/Anxiety, there is in fact an increase in r/depression (Figure \ref{fig:liwc-I}). One explanation for this is that those posting in r/Anxiety may be engaging in more discussion related to what is going in the world around them than those in r/depression, who are to a larger extent sharing their own experiences. The increase of usage of the {\sc I} pronoun is concerning because it has been shown to correlate with depression, indicating that an increase in its use could be related to worsening symptoms \cite{Rude2004}. 

Finally, we see a notable drop in the {\sc Work} category (Figure \ref{fig:liwc-WORK}), which suggests that up to this point, the stress and change associated with adapting to working from home, or worse, losing one's job, has not been a frequent topic of discussion in these forums. The drop in discussion of work is unexpected, as we believed that the economic downturn would be a significant motivator of posts as people lost jobs. We believe that this drop could be due to a decrease in work related stressors, which have been shown to cause anxiety and depression \cite{Melchior2007,Cherry1978}. It is also possible that compared to the general population, Reddit users are more likely to have jobs that can be done remotely during the pandemic, as they are more likely to have college degrees.\footnote{https://www.statista.com/statistics/517222/reddit-user-distribution-usa-education/} 

\subsubsection{Topic Analysis}
\begin{table*}[!ht]
\centering
\resizebox{\textwidth}{!}
{\begin{tabular}{@{}lcc|lcc|lcc@{}}
\toprule
\multicolumn{3}{c|}{r/Anxiety}            & \multicolumn{3}{c|}{r/depression}         & \multicolumn{3}{c}{r/SuicideWatch}       \\ \midrule
Topic & \% Outliers &    & Topic & \% Outliers &    & Topic & \% Outliers &    \\ \midrule
Transport and Daily Life\textsuperscript{*}         & 88       & {\color{red}$\downarrow$} & Family and Home\textsuperscript{*}            & 83       & {\color{customgreen}$\uparrow$} & Transport and Daily Life\textsuperscript{*}           & 70       & {\color{red}$\downarrow$}   \\
Anxiety\textsuperscript{*}           & 75       & {\color{customgreen}$\uparrow$} & Transport and Daily Life\textsuperscript{*}           & 82       & {\color{red}$\downarrow$} & Family and Home\textsuperscript{*}         & 50       & {\color{customgreen}$\uparrow$}   \\
Information Sharing\textsuperscript{*}            & 68      & {\color{customgreen}$\uparrow$} & Information Sharing\textsuperscript{*}         & 62       & {\color{customgreen}$\uparrow$} & Friends\textsuperscript{*}       & 32       & {\color{red}$\downarrow$}   \\
School\textsuperscript{*}           & 62       & {\color{red}$\downarrow$}   & Work\textsuperscript{*}         & 55       & {\color{red}$\downarrow$}   & Anxiety\textsuperscript{*}        & 28       & {\color{customgreen}$\uparrow$} \\
Family and Home\textsuperscript{*}         & 48       & {\color{customgreen}$\uparrow$} & Suicide\textsuperscript{*}         & 50       & {\color{red}$\downarrow$}   & Family and Children      & 22       & {\color{customgreen}$\uparrow$}   \\
Life and Philosophy\textsuperscript{*}        & 34       & {\color{customgreen}$\uparrow$} & Sleep and Routine\textsuperscript{*}        & 50       & {\color{red}$\downarrow$} & ``Game-over" Mentality and Swearing          & 15       & {\color{customgreen}$\uparrow$}   \\
Work\textsuperscript{*}            & 33       & {\color{red}$\downarrow$}   & Communication\textsuperscript{*}        & 48       & {\color{customgreen}$\uparrow$}   & Suicide     & 14       & {\color{red}$\downarrow$} \\ ``Game-over" Mentality and Swearing\textsuperscript{*}            & 29       & {\color{red}$\downarrow$}   & ``Game-over" Mentality and Swearing\textsuperscript{*}         & 39       & {\color{customgreen}$\uparrow$} & Worry   & 14       & {\color{customgreen}$\uparrow$} \\
Experience and Mental State\textsuperscript{*}        & 27       & {\color{customgreen}$\uparrow$}   & Medical Treatment\textsuperscript{*}           & 36       & {\color{red}$\downarrow$} &        Communication  & 13       & {\color{customgreen}$\uparrow$}   \\
Motivation\textsuperscript{*}       & 22       & {\color{red}$\downarrow$}   & Family and Children\textsuperscript{*}      & 34       & {\color{customgreen}$\uparrow$}   & People and Behavior          & 13       & {\color{customgreen}$\uparrow$} \\ \bottomrule
\end{tabular}}
\caption{Ten topics with the most outliers for r/Anxiety, r/depression, and r/SuicideWatch. Arrows mark the direction in which the mean of the outliers shifted from the predicted mean. Topics marked with * have a statistically significant percentage of outliers ($\alpha = 0.05$ before Bonferroni correction).}
\label{tab:topic-results}
\end{table*}

\begin{figure*}[!ht]
    \centering
	\begin{subfigure}{.35\textwidth}
		\centering
		\includegraphics[width=\linewidth]{legend_users_posts_comments.png}
		\label{fig:legend_users_posts_comments}
	\end{subfigure}
	
	\vspace{-10pt}
	\centering
	\begin{subfigure}[b]{.33\textwidth}
		\centering
		\includegraphics[width=\linewidth]{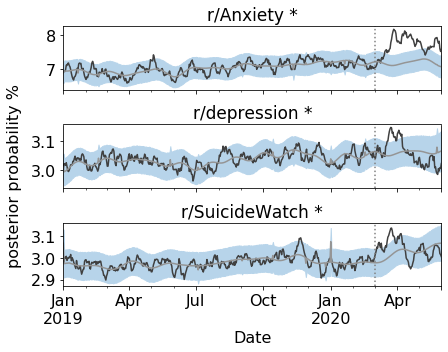}
		\caption{Anxiety topic}\label{fig:topic-anxiety}		
	\end{subfigure}%
	\begin{subfigure}[b]{.33\textwidth}
		\centering
		\includegraphics[width=\linewidth]{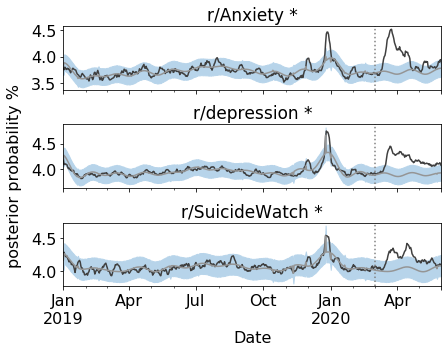}
		\caption{Family and Home topic}\label{fig:topic-family-home}
	\end{subfigure}%
	\begin{subfigure}[b]{.33\textwidth}
		\centering
		\includegraphics[width=\linewidth]{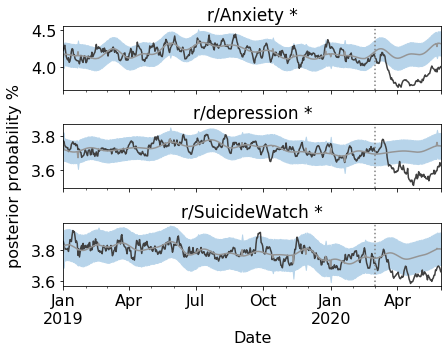}
		\caption{Transport and Daily Life topic}\label{fig:topic-transport-daily-life}
	\end{subfigure}%

	\caption{Average daily posterior probability of selected topics in posts over time. The grey line is the Prophet forecast, the shaded area is the 95\% prediction interval, and the black line is the true value. Subreddits marked with * have a statistically significant percentage of outliers ($\alpha = 0.05$ before Bonferroni correction).}\label{fig:topic-timeseries}
\end{figure*}
 We report the ten topics with the highest proportion of outliers for each subreddit during the COVID-19 period (March to May 2020) in Table~\ref{tab:topic-results}. One noteable trend is an increase in the amount of discussion related to family; we find that the {\sc Family and Home} topic increased significantly in all three subreddits and the {\sc Family and Children} topic increased significantly in r/depression. This observation is largely expected, as quarantine policies implemented to help contain COVID-19 have resulted in many people spending more time at home and with family than they had previously. Figure~\ref{fig:topic-family-home} shows how the usage of the {\sc Family and Home} topic has changed since January 2019 within each subreddit. While there are noticeable increases in all three subreddits, we a see particularly large spike in r/Anxiety starting around mid-March. Prior studies on disease outbreaks have found that uncertainty regarding the well-being of loved ones is common source of anxiety during epidemics, which may help to explain this finding \cite{Chew2020}. Another contributing factor may be the emergence of new family responsibilities, such as childcare and  home-schooling, that many people have had to take on in the face of closures caused by the pandemic.
    
Several of the results in Table~\ref{tab:topic-results} seem to reflect the disruption to normal daily life caused by COVID-19 and the resulting quarantine measures. Within all subreddits, we see a significant decrease in the {\sc Transport and Daily Life} topic (see Figure~\ref{fig:topic-transport-daily-life}), which is associated with words such ``drive," ``car," ``time," and ``day". This finding is intuitive; quarantine practices following COVID-19 have led to a large reduction in driving and other forms of transportation \cite{domonoske_adeline_2020} and, more generally, to a disruption in daily lifestyles. To the extent that these results indicate an abandonment of routine, they are  also somewhat concerning, as evidence from prior outbreaks suggests that getting back into normal routines helps to reduce loneliness and anxiety during quarantines \cite{Huremovic2019}. Mirroring the reduction of {\sc Work}-related language we observed in Section~\ref{sec:liwc-results}, we also find that there has been a significant decrease in discussion of the {\sc School} and {\sc Work} topics within the r/Anxiety and r/depression subreddits. This may indicate that these previously common sources of stress have now become secondary concerns compared to the more immediate concerns associated with COVID-19.


We observe significant changes in topics that are explicitly related to mental health. One of the most prominent trends is a significant increase in discussions of {\sc Anxiety} and its symptoms (keywords include: ``panic," ``heart," and ``chest"). 
As seen in Figure~\ref{fig:topic-anxiety}, we see a spike in {\sc Anxiety} in mid-March in all three subreddits; however, whereas we see a return to a typical level in both r/depression and r/SuicideWatch, within the r/Anxiety, {\sc Anxiety} discussion rates have remained abnormally high all the way through the end of May. These findings are aligned with existing research that has found that anxiety and the somatic symptoms associated with it are common psychological responses to epidemics \cite{Chew2020}. Further, studies of prior epidemics have found that feelings of anxiety and fear can persist even after the disease itself has been contained \cite{usher2020life}.

We find that both {\sc Information Sharing} (keywords include: ``post," ``read," ``share," ``find," and ``hope") and {\sc Communication} (keywords include: ``talk," ``call," and ``message") have become more frequent topics of discussion. This finding may be tied to the effects of social distancing measures, which have limited in-person interactions and led people to increasingly turn to digital methods of communication. These observations may also reflect a desire to seek out information related to COVID-19; individuals who experience health anxiety are more likely to exhibit online health information seeking behavior \cite{mcmullan2019relationships}. The increase in mentions of words related to social media (e.g. ``post," ``share") is somewhat worrisome; studies of disaster events have found that both more frequent social media use and exposure to conflicting information online (a widely acknowledged issue with COVID-19 \cite{kouzy2020coronavirus}) lead to higher stress levels \cite{torales2020outbreak}. However, the rise of the {\sc Information Sharing} topic, especially in it's relation to words like ``share," ``hope," and ``story," could also be indicative of a collective coping process, in which individuals come together for social support. 
As noted in Section~\ref{sec:liwc-results}, this type of coping strategy has frequently been observed during past disease outbreaks \cite{Chew2020} and may also be reflected by the increase in the usage of {\sc We} we saw for discussions in r/Anxiety.

\section{Conclusions}
In this study, we examined how COVID-19 has influenced the online behavior of individuals seeking support for mental health concerns by analyzing activity within the r/Anxiety, r/depression, and r/SuicideWatch communities on Reddit. We found substantial evidence of increases in anxiety; we observed significant increases in user activity in r/Anxiety, as well as significant increases in discussions of anxiety and the symptoms associated with it. Interestingly, we observed a decrease in activity within the r/depression and r/SuicideWatch subreddits. The literature on the impact of disease outbreaks on depression rates contains somewhat contradictory findings; we therefore believe that this is an interesting area for future work.

We also observed interesting changes in the content of discussions within each subreddit. Our results suggest that concerns related to COVID-19, such as health and family, have become more prominent discussion topics compared to other common concerns, such as work and school, which have generated less discussion since the outbreak. While our findings largely confirm the warnings offered by psychiatrists regarding the potential for COVID-19 to have an adverse effect on mental health, we also found some reason for optimism; increases in the usage of ``we” as well as the {\sc Information Sharing} topic (associated with words such as ``story" and ``hope"), suggest a heightened sense of community and shared experience, which may help individuals cope with these stressful times.





\bibliography{anthology,acl2020}

\begin{thebibliography}{39}
\expandafter\ifx\csname natexlab\endcsname\relax\def\natexlab#1{#1}\fi

\bibitem[{Ashokkumar and Pennebaker(2020)}]{pandemicprojectblog}
Ashwini Ashokkumar and James~W Pennebaker. 2020.
\newblock Turning inward during crises: How {COVID} is changing our social
  ties.
\newblock \\
  \href{https://web.archive.org/web/20200606173810/https://trackingsociallife.wordpress.com/2020/05/05/turning-inward-during-crises-how-covid-is-changing-our-social-ties/}
  {\nolinkurl{https://trackingsociallife.wordpress.com/}}.
\newblock Accessed: 2020-05-05.

\bibitem[{Baumgartner et~al.(2020)Baumgartner, Zannettou, Keegan, Squire, and
  Blackburn}]{Baumgartner2020}
Jason Baumgartner, Savvas Zannettou, Brian Keegan, Megan Squire, and Jeremy
  Blackburn. 2020.
\newblock \href {http://arxiv.org/abs/2001.08435} {{The Pushshift Reddit
  Dataset}}.

\bibitem[{Blei et~al.(2003)Blei, Ng, and Jordan}]{blei2003latent}
David~M Blei, Andrew~Y Ng, and Michael~I Jordan. 2003.
\newblock Latent dirichlet allocation.
\newblock \emph{Journal of machine Learning research}, 3(Jan):993--1022.

\bibitem[{Chakravorti et~al.(2018)Chakravorti, Law, Gemmell, and
  Raicu}]{chakravorti2018detecting}
Dante Chakravorti, Kathleen Law, Jonathan Gemmell, and Daniela Raicu. 2018.
\newblock Detecting and characterizing trends in online mental health
  discussions.
\newblock In \emph{2018 IEEE International Conference on Data Mining Workshops
  (ICDMW)}, pages 697--706. IEEE.

\bibitem[{Chancellor and De~Choudhury(2020)}]{chancellor2020methods}
Stevie Chancellor and Munmun De~Choudhury. 2020.
\newblock Methods in predictive techniques for mental health status on social
  media: a critical review.
\newblock \emph{NPJ digital medicine}, 3(1):1--11.

\bibitem[{Cherry(1978)}]{Cherry1978}
Nicola Cherry. 1978.
\newblock \href {https://doi.org/10.1111/j.2044-8325.1978.tb00422.x} {{Stress,
  anxiety and work: A longitudinal study.}}
\newblock \emph{Journal of Occupational Psychology}, 51(3):259--270.

\bibitem[{Chew et~al.(2020)Chew, Wei, Vasoo, Chua, and Sim}]{Chew2020}
QH~Chew, KC~Wei, Shawn Vasoo, HC~Chua, and Kang Sim. 2020.
\newblock \href {https://doi.org/10.11622/smedj.2020046} {{Narrative synthesis
  of psychological and coping responses towards emerging infectious disease
  outbreaks in the general population: practical considerations for the
  COVID-19 pandemic}}.
\newblock \emph{Singapore Medical Journal}, (April):1--31.

\bibitem[{De~Choudhury and De(2014)}]{de2014mental}
Munmun De~Choudhury and Sushovan De. 2014.
\newblock Mental health discourse on reddit: Self-disclosure, social support,
  and anonymity.
\newblock In \emph{Eighth international AAAI conference on weblogs and social
  media}.

\bibitem[{De~Choudhury et~al.(2016)De~Choudhury, Kiciman, Dredze, Coppersmith,
  and Kumar}]{de2016discovering}
Munmun De~Choudhury, Emre Kiciman, Mark Dredze, Glen Coppersmith, and Mrinal
  Kumar. 2016.
\newblock Discovering shifts to suicidal ideation from mental health content in
  social media.
\newblock In \emph{Proceedings of the 2016 CHI conference on human factors in
  computing systems}, pages 2098--2110.

\bibitem[{Domonoske and Adeline(2020)}]{domonoske_adeline_2020}
Camila Domonoske and Stephanie Adeline. 2020.
\newblock \href
  {https://web.archive.org/web/20200623073738/https://www.npr.org/sections/coronavirus-live-updates/2020/05/06/851001762/the-pandemic-emptied-american-roads-but-driving-is-picking-back-up}
  {The pandemic emptied american roads. but driving is picking back up}.
\newblock \emph{NPR}.

\bibitem[{Feldhege et~al.(2020)Feldhege, Moessner, and
  Bauer}]{feldhege2020says}
Johannes Feldhege, Markus Moessner, and Stephanie Bauer. 2020.
\newblock Who says what? content and participation characteristics in an online
  depression community.
\newblock \emph{Journal of Affective Disorders}, 263:521--527.

\bibitem[{Greenberg et~al.(2020)Greenberg, Docherty, Gnanapragasam, and
  Wessely}]{greenberg2020managing}
Neil Greenberg, Mary Docherty, Sam Gnanapragasam, and Simon Wessely. 2020.
\newblock Managing mental health challenges faced by healthcare workers during
  covid-19 pandemic.
\newblock \emph{bmj}, 368.

\bibitem[{Huang et~al.(2020)Huang, Jamison, Broniatowski, Quinn, and
  Dredze}]{huang_xiaolei_2020_3735015}
Xiaolei Huang, Amelia Jamison, David Broniatowski, Sandra Quinn, and Mark
  Dredze. 2020.
\newblock \href {https://doi.org/10.5281/zenodo.3897727} {{Coronavirus Twitter
  Data: A collection of COVID-19 tweets with automated annotations}}.
\newblock Http://twitterdata.covid19dataresources.org/index.

\bibitem[{Huremovi{\'{c}}(2019)}]{Huremovic2019}
Damir Huremovi{\'{c}}. 2019.
\newblock \href {http://link.springer.com/10.1007/978-3-030-15346-5{\_}8}
  {\emph{{Psychiatry of Pandemics: A Mental Health Response to Infection
  Outbreak}}}.
\newblock Springer International Publishing, Cham.

\bibitem[{Jacobson et~al.(2020)Jacobson, Lekkas, Price, Heinz, Song,
  O’Malley, and Barr}]{jacobson2020flattening}
Nicholas~C Jacobson, Damien Lekkas, George Price, Michael~V Heinz, Minkeun
  Song, A~James O’Malley, and Paul~J Barr. 2020.
\newblock Flattening the mental health curve: {COVID}-19 stay-at-home orders
  are associated with alterations in mental health search behavior in the
  united states.
\newblock \emph{JMIR mental health}, 7(6):e19347.

\bibitem[{Kouzy et~al.(2020)Kouzy, Abi~Jaoude, Kraitem, El~Alam, Karam, Adib,
  Zarka, Traboulsi, Akl, and Baddour}]{kouzy2020coronavirus}
Ramez Kouzy, Joseph Abi~Jaoude, Afif Kraitem, Molly~B El~Alam, Basil Karam,
  Elio Adib, Jabra Zarka, Cindy Traboulsi, Elie~W Akl, and Khalil Baddour.
  2020.
\newblock Coronavirus goes viral: quantifying the {COVID}-19 misinformation
  epidemic on twitter.
\newblock \emph{Cureus}, 12(3).

\bibitem[{Kumar et~al.(2015)Kumar, Dredze, Coppersmith, and {De
  Choudhury}}]{Kumar2015}
Mrinal Kumar, Mark Dredze, Glen Coppersmith, and Munmun {De Choudhury}. 2015.
\newblock \href {https://doi.org/10.1145/2700171.2791026} {{Detecting Changes
  in Suicide Content Manifested in Social Media Following Celebrity Suicides}}.
\newblock In \emph{Proceedings of the 26th ACM Conference on Hypertext {\&}
  Social Media - HT '15}, volume 176, pages 85--94, New York, New York, USA.
  ACM Press.

\bibitem[{Li et~al.(2020)Li, Wang, Xue, Zhao, and Zhu}]{li2020impact}
Sijia Li, Yilin Wang, Jia Xue, Nan Zhao, and Tingshao Zhu. 2020.
\newblock The impact of {COVID}-19 epidemic declaration on psychological
  consequences: a study on active weibo users.
\newblock \emph{International journal of environmental research and public
  health}, 17(6):2032.

\bibitem[{Li et~al.(2018)Li, Mihalcea, and Wilson}]{li2018text}
Yaoyiran Li, Rada Mihalcea, and Steven~R Wilson. 2018.
\newblock Text-based detection and understanding of changes in mental health.
\newblock In \emph{International Conference on Social Informatics}, pages
  176--188. Springer.

\bibitem[{McCallum(2002)}]{mallet}
Andrew~Kachites McCallum. 2002.
\newblock \href {http://mallet.cs.umass.edu} {Mallet: A machine learning for
  language toolkit}.

\bibitem[{McMullan et~al.(2019)McMullan, Berle, Arn{\'a}ez, and
  Starcevic}]{mcmullan2019relationships}
Ryan~D McMullan, David Berle, Sandra Arn{\'a}ez, and Vladan Starcevic. 2019.
\newblock The relationships between health anxiety, online health information
  seeking, and cyberchondria: Systematic review and meta-analysis.
\newblock \emph{Journal of affective disorders}, 245:270--278.

\bibitem[{Melchior et~al.(2007)Melchior, Caspi, Milne, Danese, Poulton, and
  Moffitt}]{Melchior2007}
Maria Melchior, Avshalom Caspi, Barry~J Milne, Andrea Danese, Richie Poulton,
  and Terrie~E Moffitt. 2007.
\newblock \href {https://doi.org/10.1017/S0033291707000414} {{Work stress
  precipitates depression and anxiety in young, working women and men}}.
\newblock \emph{Psychological medicine}, 37(8):1119--1129.

\bibitem[{Mimno et~al.(2011)Mimno, Wallach, Talley, Leenders, and
  McCallum}]{mimno2011optimizing}
David Mimno, Hanna Wallach, Edmund Talley, Miriam Leenders, and Andrew
  McCallum. 2011.
\newblock Optimizing semantic coherence in topic models.
\newblock In \emph{Proceedings of the 2011 Conference on Empirical Methods in
  Natural Language Processing}, pages 262--272.

\bibitem[{Nandi et~al.(2009)Nandi, Tracy, Beard, Vlahov, and
  Galea}]{nandi2009patterns}
Arijit Nandi, Melissa Tracy, John~R Beard, David Vlahov, and Sandro Galea.
  2009.
\newblock Patterns and predictors of trajectories of depression after an urban
  disaster.
\newblock \emph{Annals of epidemiology}, 19(11):761--770.

\bibitem[{Pavalanathan and De~Choudhury(2015)}]{pavalanathan2015identity}
Umashanthi Pavalanathan and Munmun De~Choudhury. 2015.
\newblock Identity management and mental health discourse in social media.
\newblock In \emph{Proceedings of the 24th International Conference on World
  Wide Web}, pages 315--321.

\bibitem[{Pennebaker and Harber(1993)}]{Pennebaker93Social}
James Pennebaker and Kent Harber. 1993.
\newblock A social stage model of collective coping: The loma prieta earthquake
  and the persian gulf war.
\newblock \emph{Journal of Social Issues}, 49.

\bibitem[{Pennebaker et~al.(2015)Pennebaker, Boyd, Jordan, and
  Blackburn}]{pennebaker2015development}
James~W Pennebaker, Ryan~L Boyd, Kayla Jordan, and Kate Blackburn. 2015.
\newblock The development and psychometric properties of liwc2015.
\newblock Technical report.

\bibitem[{Qiu et~al.(2020)Qiu, Shen, Zhao, Wang, Xie, and
  Xu}]{qiu2020nationwide}
Jianyin Qiu, Bin Shen, Min Zhao, Zhen Wang, Bin Xie, and Yifeng Xu. 2020.
\newblock A nationwide survey of psychological distress among chinese people in
  the {COVID}-19 epidemic: implications and policy recommendations.
\newblock \emph{General psychiatry}, 33(2).

\bibitem[{Roy et~al.(2020)Roy, Tripathy, Kar, Sharma, Verma, and
  Kaushal}]{roy2020study}
Deblina Roy, Sarvodaya Tripathy, Sujita~Kumar Kar, Nivedita Sharma,
  Sudhir~Kumar Verma, and Vikas Kaushal. 2020.
\newblock Study of knowledge, attitude, anxiety \& perceived mental healthcare
  need in indian population during {COVID}-19 pandemic.
\newblock \emph{Asian Journal of Psychiatry}, page 102083.

\bibitem[{Rude et~al.(2004)Rude, Gortner, and Pennebaker}]{Rude2004}
Stephanie~S. Rude, Eva~Maria Gortner, and James~W. Pennebaker. 2004.
\newblock \href {https://doi.org/10.1080/02699930441000030} {{Language use of
  depressed and depression-vulnerable college students}}.
\newblock \emph{Cognition and Emotion}, 18(8):1121--1133.

\bibitem[{Taylor and Letham(2018)}]{Taylor2018}
Sean~J. Taylor and Benjamin Letham. 2018.
\newblock \href {https://doi.org/10.1080/00031305.2017.1380080} {{Forecasting
  at Scale}}.
\newblock \emph{The American Statistician}, 72(1):37--45.

\bibitem[{Torales et~al.(2020)Torales, O’Higgins, Castaldelli-Maia, and
  Ventriglio}]{torales2020outbreak}
Julio Torales, Marcelo O’Higgins, Jo{\~a}o~Mauricio Castaldelli-Maia, and
  Antonio Ventriglio. 2020.
\newblock The outbreak of {COVID}-19 coronavirus and its impact on global
  mental health.
\newblock \emph{International Journal of Social Psychiatry}, page
  0020764020915212.

\bibitem[{Usher et~al.(2020)Usher, Bhullar, and Jackson}]{usher2020life}
Kim Usher, Navjot Bhullar, and Debra Jackson. 2020.
\newblock Life in the pandemic: Social isolation and mental health.
\newblock \emph{Journal of Clinical Nursing}.

\bibitem[{Wang et~al.(2020)Wang, Pan, Wan, Tan, Xu, Ho, and
  Ho}]{wang2020immediate}
Cuiyan Wang, Riyu Pan, Xiaoyang Wan, Yilin Tan, Linkang Xu, Cyrus~S Ho, and
  Roger~C Ho. 2020.
\newblock Immediate psychological responses and associated factors during the
  initial stage of the 2019 coronavirus disease ({COVID}-19) epidemic among the
  general population in china.
\newblock \emph{International journal of environmental research and public
  health}, 17(5):1729.

\bibitem[{Wolohan(2020)}]{wolohan2020estimating}
JT~Wolohan. 2020.
\newblock Estimating the effect of covid-19 on mental health: Linguistic
  indicators of depression during a global pandemic.
\newblock In \emph{ACL 2020 Workshop on Natural Language Processing for
  COVID-19}. Association for Computational Linguistics.

\bibitem[{Yao et~al.(2020)Yao, Chen, and Xu}]{yao2020patients}
Hao Yao, Jian-Hua Chen, and Yi-Feng Xu. 2020.
\newblock Patients with mental health disorders in the covid-19 epidemic.
\newblock \emph{The Lancet Psychiatry}, 7(4):e21.

\bibitem[{Zhang and Ma(2020)}]{zhang2020impact}
Yingfei Zhang and Zheng~Feei Ma. 2020.
\newblock Impact of the {COVID}-19 pandemic on mental health and quality of
  life among local residents in liaoning province, china: A cross-sectional
  study.
\newblock \emph{International journal of environmental research and public
  health}, 17(7):2381.

\bibitem[{Zwiers and Von~Storch(1995)}]{zwiers1995taking}
Francis~W Zwiers and Hans Von~Storch. 1995.
\newblock Taking serial correlation into account in tests of the mean.
\newblock \emph{Journal of Climate}, 8(2):336--351.

\bibitem[{Şahin et~al.(2020)Şahin, Tasci, and Yan}]{Sahin2020}
Ayşeg{\"{u}}l Şahin, Murat Tasci, and Jin Yan. 2020.
\newblock \href {https://doi.org/10.26509/frbc-ec-202009} {{The Unemployment
  Cost of COVID-19: How High and How Long?}}
\newblock \emph{Economic Commentary (Federal Reserve Bank of Cleveland)}, pages
  1--7.

\end{thebibliography}
\bibliographystyle{acl_natbib}

\appendix

\clearpage

 \section{Topics identified by LDA Model}\label{sec:topics-keywords}
 Table \ref{tab:topics-keywords} shows the topics identified by the LDA model.\\

\begin{minipage}[b]{\textwidth}
{\small
\begin{tabularx}{\textwidth}{p{0.3\textwidth}p{0.7\textwidth}}
\toprule
       Topic Label & High Probability Words \\ \midrule
 School &  school, year, college, high, class, fail, parent, study, grade, start \\[1.5mm]
 Relationships & love, relationship, girl, guy, good, girlfriend, break, date,  meet, find \\[1.5mm]
 Experience and Mental State & experience, situation, mind, part, brain, lead, state, feeling,  sense, learn \\[1.5mm]
 Communication & talk, call, time, phone, send, text, give, back, speak, message \\[1.5mm]
 People and Behavior & people, make, person, care, thing, understand, problem, wrong,  act, attention \\[1.5mm]
 Feelings & happy, tired, cry, anymore, sad, make, hurt, depressed, stop,  feeling \\[1.5mm]
 “Game-over” Mentality and Swearing & hate, fuck, shit, fucking, die, wanna, stupid, kill, literally, idk \\[1.5mm]
 Transport and Daily Life & drive, time, back, car, drink, start, walk, home, run, day \\[1.5mm]
 Time &  year, month, start, time, back, ago, day, week, past, couple \\[1.5mm]
 Worry & thought, mind, fear, worry, head, afraid, scared, scare, stop, happen \\[1.5mm]
 Friends & friend, talk, people, good, social, play, make, close, hang, group \\[1.5mm]
 Anxiety & anxiety, attack, panic, anxious, heart, symptom, calm, chest, experience, stress \\[1.5mm]
 Medical Treatment & anxiety, medication, doctor, therapy, med, therapist, experience, week, mg, work \\[1.5mm]
 Body and Food & eat, body, eye, face, hand, head, sit, food, weight, walk \\[1.5mm]
 - & bad, thing, make, time, lot, happen, pretty, good, stuff, kind \\[1.5mm]
 Life and Philosophy & life, world, hope, exist, dream, human, live, pain, love, real \\[1.5mm]
 Depression and Mental Illness & depression, mental, issue, health, problem, struggle, deal, bad, suffer, year \\[1.5mm]
 Life Purpose & life, live, end, anymore, point, family, reason, care, worth, future \\[1.5mm]
 Motivation & thing, time, make, good, find, hard, work, enjoy, change, motivation \\[1.5mm]
 Work & work, job, money, pay, quit, find, afford, interview, company, month \\[1.5mm]
 Family and Children & year, family, mother, kid, parent, child, life, father, young, age \\[1.5mm]
 Information Sharing & post, read, write, find, hope, give, share, story, reddit, long \\[1.5mm]
 Family and Home & leave, mom, home, move, house, dad, family, live, parent, stay \\[1.5mm]
 Sleep and Routine & day, sleep, night, hour, wake, today, bed, morning, work, week \\[1.5mm]
 Suicide & kill, die, suicide, pain, suicidal, end, attempt, cut, plan, dead \\[1.5mm]

 \bottomrule
\end{tabularx}
}
\captionof{table}{Topics identified by the LDA topic model. For each topic, we provide a summary label and the ten most probable words. We omit labels for topics whose keywords did not have a clear interpretation.} \label{tab:topics-keywords}
\end{minipage}

\clearpage

\section{Statistical Significance Test}\label{sec:sig-testing}
We apply a one-sample proportion test to assess whether the proportion of observations outside of the prediction interval in the post-COVID period is significantly greater than 5\%. This test assumes that the observations are independent; however, we find that there is order-1 autocorrelation in our data. We therefore apply a correction for order-1 autocorrelation \cite{zwiers1995taking} when computing the z-test statistic. The corrected test statistic is: 

\begin{equation}
    z = \frac{\hat{p} - p_0}{\sqrt{p_0(1 - p_0)(1 + r)/n(1-r)}}
\end{equation}

\noindent where $\hat{p}$ is the proportion of observations outside of the prediction interval in the post-COVID period, $p_0 = .05$, $n$ is the number of observations in the post-COVID period, and $r$ is the lag-1 correlation coefficient of the pre-COVID data.

We use a Bonferroni correction when determining statistical significance for our discussion content metrics, as we ran almost 300 tests. $M = 294$, which is the number of LIWC categories and topics, multiplied by the number of subreddits. Our corrected $\alpha = 0.05/294 = \num{1.7e-5}$.

\end{document}